\documentclass[10pt, a4paper]{article}

\usepackage{lrec-coling2024} 

\usepackage{times}
\usepackage{latexsym}
\usepackage{amssymb}
\usepackage{amsmath}
\usepackage{adjustbox}
\usepackage{makecell}
\usepackage{multirow}
\usepackage{booktabs}
\usepackage[ruled,linesnumbered]{algorithm2e}

\newcommand{\isep}{\mathrel{{.}\,{.}}\nobreak}

\title{MemoryPrompt: A Light Wrapper to Improve Context Tracking in Pre-trained Language Models}

\name{Nathanaël Carraz Rakotonirina$^1$, Marco Baroni$^{1,2}$} 

\address{$^1$Universitat Pompeu Fabra, Barcelona, Spain\\
         $^2$ICREA, Barcelona, Spain \\
         \{nathanael.rakotonirina, marco.baroni\}@upf.edu\\}

\abstract{
Transformer-based language models (LMs) track contextual information through large, hard-coded input windows. We introduce \textit{MemoryPrompt}, a leaner approach in which the LM is complemented by a small auxiliary recurrent network that passes information to the LM by prefixing its regular input with a sequence of vectors, akin to \textit{soft prompts}, without requiring LM finetuning. Tested on a task designed to probe a LM's ability to keep track of multiple fact updates, a MemoryPrompt-augmented LM outperforms much larger LMs that have access to the full input history. We also test MemoryPrompt on a long-distance dialogue dataset, where its performance is comparable to that of a model conditioned on the entire conversation history. In both experiments we also observe that, unlike full-finetuning approaches, MemoryPrompt does not suffer from catastrophic forgetting when adapted to new tasks, thus not disrupting the generalist capabilities of the underlying LM.
 \\ \newline  }

\begin{document}

\maketitleabstract

\begin{figure*}
    \centering
    \includegraphics[width=0.9\textwidth]{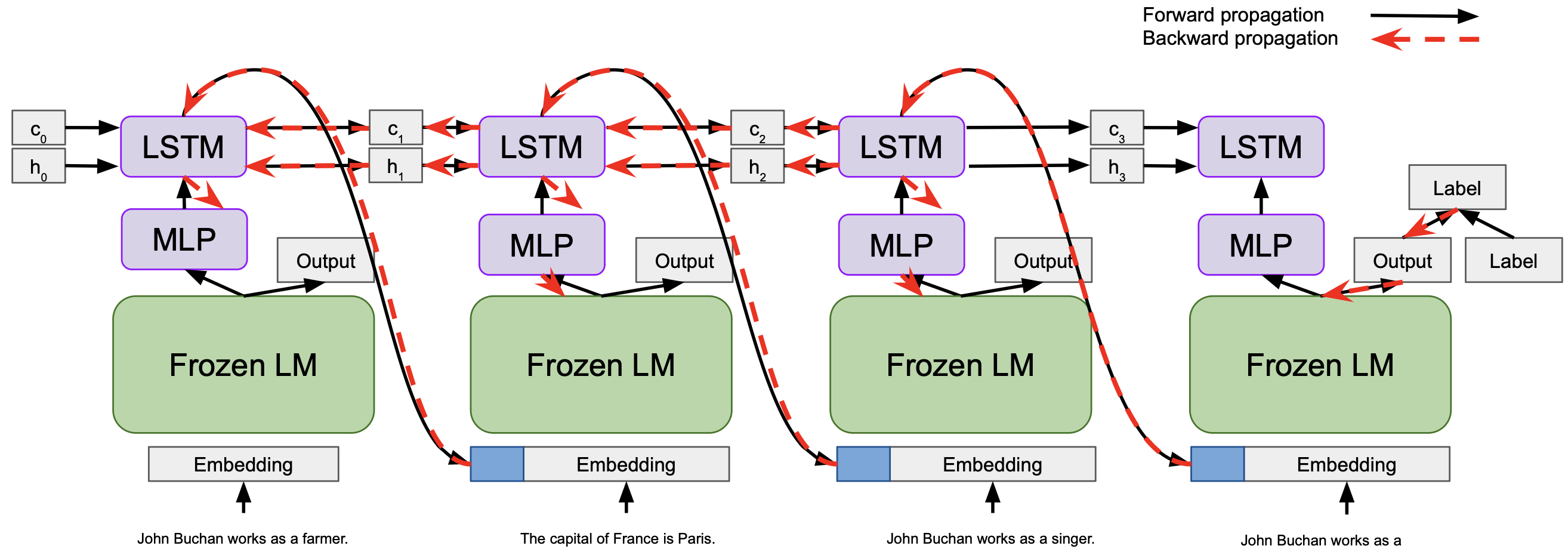}
    \caption{Unfolded graph of MemoryPrompt at training time. The input is divided into segments and, for each segment, the augmented system produces both the LM output and the memory vectors (blue) which are concatenated to the embeddings of the next segment. %
    }
    \label{fig:graph}
\end{figure*}

\section{Introduction}

Transformer-based language models (LMs) can only track user-provided contextual information if it fits into their context window. The brute-force solution of using a huge window suffers obvious problems of scale. While there is promising work in the direction of making long contexts more computationally efficient \citep{Dai:etal:2019,Beltagy:etal:2020,chen2023extending}, we introduce MemoryPrompt, a cheaper and complementary solution that augments a pre-trained LM with a smaller auxiliary recurrent network, trained to carry context-relevant information through time. Inspired by work on ``soft prompting'' \citep{Lester:etal:2021,Liu:etal:2021,Zhong:etal:2021}, this information is passed to the LM at each time step as a continuous token prefixed to its regular input. Our method does not  architecturally alter or finetune the underlying pre-trained LM, and it can thus leverage any available pre-implemented model. Memory augmentation is moreover unobtrusive, in the sense that it does not greatly affect the underlying LM behaviour in standard next-token prediction tasks, and thus the augmented model can successfully manage contextually-driven updates, while preserving the knowledge encoded in the underlying LM.\footnote{The code and data to reproduce our analysis is available at \url{https://github.com/ncarraz/MemoryPrompt}.} 

\section{Related work}

Methods to enhance a sequence processing network with an external differentiable memory have been explored since the comeback of neural networks during the last decade \citep[e.g.,][]{Joulin:Mikolov:2015,Sukhbaatar:etal:2015,Graves:etal:2016}. 

After the advent of the Transformer \citep{Vaswani:etal:2017}, much work on long-range memory has focused on how to make its attention mechanisms more efficient, in order to handle a larger span \citep[e.g.,][]{Dai:etal:2019,Beltagy:etal:2020}. Similarly to us, \citet{Fan:etal:2020} and \citet{hutchins2022block} introduce a recurrence to allow the Transformer to carry information through time steps, but they do it by modifying the core Transformer architecture. 

The idea of using a differentiable external memory has also made a comeback in the context of Transformer-based language models. A recent example is the Memformer architecture of \citet{Wu:etal:2022}. Like in our approach, Memformer has an external memory component that interacts with a Transformer via reading and writing operations. However, this approach demands architectural changes to the Transformer, special reading/writing operations and end-to-end training of the memory and the underlying Transformer, preventing its use with pre-trained LMs.

The closest approach to ours is the recently introduced Recurrent Memory Transformer (RMT) model of \citet{Bulatov:etal:2022} \citep[see also][]{Bulatov:etal:2023}. They divide the input into segments, and add the same real-valued memory vectors both at the beginning and at the end of the segment. Then, the activations of the memory vectors added at the end of the segment form the memory vectors of the next segment in a recurrent manner. The key difference between RMT and our approach is that we use a lightweight module to produce the memory vectors and keep our base model frozen, updating only the parameters of the module. As we will show below, this is crucial to prevent catastrophic forgetting of the knowledge that was originally encoded in the base model.

Our idea of passing information to the Transformer in the form of continuous tokens fed to its standard input comes from the literature on soft prompting \citep[e.g.,][]{Lester:etal:2021,Li:Liang:2021,Liu:etal:2021,Zhong:etal:2021}, where sequences of vectors living in the target LM's embedding space are prefixed to task-specific inputs, to implicitly adapt the model to the task without fine-tuning. We  extend continuous prompts to a setup where, instead of a fixed prefix, a dynamic one must be generated at each step, in order to carry constantly updated information.

\section{The MemoryPrompt model}
We augment an autoregressive language model with a recurrent memory module, allowing it to extend its effective context-length and to keep track of information updated through time. 

The input is divided into segments that are sequentially processed by the model. For each segment, the contextual representation of the last token is fed as input to the memory module which is composed of an MLP followed by an LSTM \citep{Hochreiter:Schmidhuber:1997}. The output of the memory module is a series of memory vectors $\mathbf{P} \in \mathbb{R}^{m\times e}$, where $e$ is the word embedding space and $m$ is the number of vectors. 

The augmented system is trained end-to-end with backpropagation through time (see Figure \ref{fig:graph}), but only the parameters of the memory module are updated: the LM components are kept frozen. Algorithm \ref{alg:forward_prop} illustrates the forward process of MemoryPrompt in detail.

\begin{algorithm}[tb]
\begin{footnotesize}
\caption{Forward flow of MemoryPrompt for a single input}
\label{alg:forward_prop}
\SetKwInOut{Hyperparameters}{Hyperparameters}
\Hyperparameters{number of memory vectors $m$ \\ number of blocks $B$} 
Initialize the hidden state $\mathbf{h}$ and cell state $\mathbf{c}$ \\
Segment the input into $B$ blocks of $n$ tokens $\mathbf{X}=\{\mathbf{X_1},...,\mathbf{X_B}\}$ \\
Embed the input $\mathbf{X}$ to get  $\mathbf{E}=\{\mathbf{E_1},...,\mathbf{E_B}\}$ where $\mathbf{E_i} \in \mathbb{R}^{n \times e}$\\
Get the activations before the last linear layer $\mathbf{A} = \textit{LM}_\phi(\mathbf{E_1})$ where $\mathbf{A} \in \mathbb{R}^{n \times e}$\\
Compute the memory vector from the activations of the last token $\mathbf{p}, \mathbf{h}, \mathbf{c} =\textit{LSTM}_\theta(\textit{MLP}_\theta(\mathbf{A}[-1,:]), \mathbf{h}, \mathbf{c})$ where $\mathbf{p}$ is a vector of length ${m\times e}$  \\
Reshape $\mathbf{p}$ into the matrix $\mathbf{P} \in \mathbb{R}^{m\times e}$ \\
Get the probabilities over the vocabulary using the last linear layer     $\mathbf{O} = \textit{Softmax}(\mathbf{A}\mathbf{W}_\phi)$ where $\mathbf{W}_\phi \in \mathbb{R}^{e\times V}$ \\
\For{$b=2$ \KwTo $B$}{
    Concatenate the memory vectors to the embedding before feeding to the model 
    $\mathbf{A} = \textit{LM}_\phi([\mathbf{P};\mathbf{E_b}])$ where $[\mathbf{P};\mathbf{E_b}] \in \mathbb{R}^{(m+n)\times e}$ \\
    Compute the next memory vector, hidden state and cell state
    $\mathbf{p}, \mathbf{h}, \mathbf{c} =\textit{LSTM}_\theta(\textit{MLP}_\theta(\mathbf{A}[-1,:]), \mathbf{h}, \mathbf{c})$ \\
    Reshape $\mathbf{p}$ into the matrix $\mathbf{P}$  \\
    Get the probabilities over the vocabulary $\mathbf{O} =\textit{Softmax}(\mathbf{A}\mathbf{W}_\phi)$ \\
}
\end{footnotesize}
\end{algorithm} 

\section{Experimental setup} 

\subsection{Datasets}
\paragraph{Fact updating} We simulate a scenario in which the model is exposed to realistic fact updates that it needs to track. We use sequences of facts gathered from the version of TREx \citep{elsahar2018t} curated by \citet{Elazar:etal:2021}. Each fact is represented as a triple $\langle subject,relation,object \rangle$---for example, $\langle \textit{Antim Peak}, \textit{continent}, \textit{Antarctica}\rangle$. As this  example shows, not all facts can be plausibly updated. We thus identified 3 \textit{mutable} relations, for which an update would be credible, namely
\textit{employer} (\textit{Paul Allen}/\textit{Microsoft}), \textit{position held} (\textit{Otto Suhr}/\textit{mayor}) and \textit{work location} (\textit{Lucio Fontana}/\textit{Milan}). These were updated by randomly selecting other objects from the same TREx relation pool (e.g., \textit{Paul Allen}'s employer might be randomly updated to \textit{Apple} or \textit{BBC}). We used the remaining 36 relations to generate \textit{stable} facts that are not updated. Each fact is instantiated as a natural-language statement using a template from  \citet{Elazar:etal:2021} (e.g.,  
the template for the \textit{employer} relation is \textit{[subject] works for [object]}).

\begin{figure*}[tb]
    \centering
    \includegraphics[width=\textwidth]{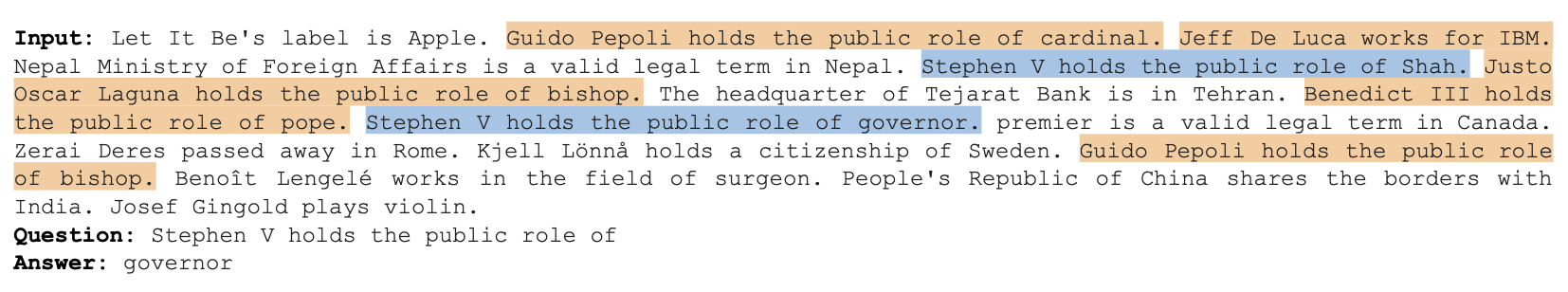}
    \caption{Sequence from the short-fd dataset (see Table \ref{tab:dataset_config}). Non-highlighted text contains stable facts. The pivot (in blue) is the mutable fact to track, and the final answer is the most up-to-date object of the pivot. Distractors (in orange) are mutable facts distinct form the pivot, which might belong to the same relation and might be updated (here, the object associated with \textit{Guido Pepoli} is updated from \textit{cardinal} to \textit{bishop}).}
    \label{fig:sequence}
\end{figure*} 

A \textit{sequence} in the dataset is composed of multiple statements. A statement can either introduce a new fact or update a previously introduced mutable fact. The model is asked to predict the most up-to-date object of a single mutable fact called the \textit{pivot} fact. The other mutable facts in the sequence, that can also be updated, are called \textit{distractors}. Note that a statement can pertain to a pivot fact, to another mutable fact (a distractor), or to a stable fact. Nothing distinguishes pivots from distractors, so a model must be able to track updates to both fact types.

An example sequence is shown in Figure \ref{fig:sequence}. We generate different versions of the fact-updating dataset as shown in Table \ref{tab:dataset_config}. Each dataset contains 26,892 sequences for training, 150 for validation and 346 for testing. The number of examples in the validation and test sets is limited by the number of mutable facts in TREx. For training, we could generate an arbitrarily large number of sequences by randomly selecting objects. However, performing this augmentation on validation or test data does not affect the final performance; hence the smaller size of the latter splits.

Note that, while the datasets are automatically generated, they contain plausible statements and  updates expressed in natural language, and one could easily imagine concrete scenarios in which similar fact streams arise (e.g., processing a stream of messages from a news agency).

\paragraph{Multi-Session Chat (MSC)} We also test MemoryPrompt in the less-constrained scenario of long-distance language modeling on the MSC dataset \citep{xu2021beyond}. This is a long-term conversation dataset which consists of multi-session crowd-worker chats, where the speakers might refer to their whole shared history. The dataset contains 35,880 training dialogues, 5,000 validation examples and 5,008 test cases.

\begin{table*}[tb]
\centering
\begin{tabular}{lccc}
\toprule
\textbf{Dataset}        & \multicolumn{1}{l}{\textbf{\begin{tabular}[c]{@{}l@{}}no. \\ statements\end{tabular}}} & \multicolumn{1}{l}{\textbf{\begin{tabular}[c]{@{}l@{}}no. \\ distractors\end{tabular}}} & \multicolumn{1}{l}{\textbf{\begin{tabular}[c]{@{}l@{}}no. \\ updates\end{tabular}}} \\
\midrule
short no distractors (nd)  & $[10\isep30]$                                     & $0$                                            & $[0\isep4]$                                        \\
short few distractors (fd)                  & $[10\isep30]$                                      & $[3\isep7]$                                           & $[0\isep4]$                                         \\
long no distractors (nd)  & $[150\isep190]$                                     & $0$                                            & $[0\isep9]$                                         \\
long few distractors (fd)    & $[150\isep190]$                                     & $[3\isep10]$                                          & $[0\isep9]$                                         \\
long many distractors (md) & $[150\isep190]$                                     & $[25\isep50]$                                         & $[0\isep9]$                                         \\
long many updates (mu)    & $[150\isep190]$                                     & $[3\isep10]$                                           & $[0\isep49]$     \\    
\bottomrule
\end{tabular}
\caption{Different fact-updating dataset configurations. Datasets are generated using three parameters whose values are specified by ranges: the number of statements, the number of distractors and the number of updates to the pivot fact. When generating a sequence, the value of each parameter is  randomly selected within the corresponding range. The positions of the pivot and the distractors are also randomly selected.
} 
\label{tab:dataset_config}
\end{table*}

\subsection{Models}
The MemoryPrompt memory module consists of a 1-layer 1024-dimensional MLP followed by a 1-layer LSTM. We use 5 memory vectors. We use LMs from the OPT family \citep{zhang2022opt}.  For reference, our MemoryPrompt configuration adds about 10\% new parameters to OPT-350M. Since the number of parameters of the memory module only depends on the embedding dimension of the base model, this ratio decreases as the base model gets larger. Due to computational limitations, and because we are interested in external memories as an alternative to scaling up models, we only applied MemoryPrompt and RMT to OPT-125M and OPT-350M. 

We compare the following setups. The \textbf{LM-full-context} setting involves an OPT LM that takes a whole fact-updating dataset sequence or MSC conversation history as input. In the \textbf{LM + MemoryPrompt} setup, the input sequence is segmented into blocks that are separately processed by the model. Each block contains 5 facts for the short fact-updating datasets, 10 for the long ones, and a turn for MSC.  
Only the parameters of the memory module are updated. 
\textbf{LM-finetuned + RMT} is our re-implementation of RMT, the memory-based approach of \citet{Bulatov:etal:2022}, which adds the same memory vectors both at the beginning and at the end of a segment when, as in our case, it is applied to decoder-only models. The activations of the  memory vectors added at the end of the segment form the vectors for the next segment. The LM is finetuned to produce those vectors without resorting to a separate module. The input and the number of memory vectors are the same as for MemoryPrompt. 

\subsection{Training setup}
\label{sec:training}

We train the models using the AdamW optimizer \citep{loshchilov2017decoupled} with a learning rate of 7e-5 (7e-6 for the MSC dataset) and weight decay of 0.1, without hyperparameter tuning.  
We follow a linear learning rate schedule for MSC, warming up from 0 to the maximum learning rate over the first 12,000 steps. We clip gradient norms at 1.0 to avoid exploding gradients for MSC. Moreover, we apply L2 regularization to the memory vectors, as we informally observed this to mitigate catastrophic forgetting.

In the fact-updating experiments, during training and evaluation we provide the models with demonstrations using facts from the same relation as the pivot. MemoryPrompt is robust to the number of demonstrations. In our experiments, we use 4 examples. Moreover, following \citet{Bulatov:etal:2023}, we use curriculum learning when training on longer sequences for better performance and faster convergence. The model is first trained on the short fact-updating dataset until convergence. Then, we use this as the starting point for a long dataset.

Making RMT converge on the fact-updating dataset was not trivial. Unlike MemoryPrompt, which directly converged on short fact-updating datasets, RMT required a more elaborate curriculum learning process. We first had to train it on very short sequences whose length was gradually increased upon convergence. Moreover, RMT is more prone to exploding and vanishing gradients \citep{hochreiter1998vanishing, bengio1994learning}. Training RMT is even more difficult when the model is larger. We failed to make OPT-350M-finetuned + RMT converge on any of the datasets due to this problem.

\section{Results}

\begin{table*}[tb]
\centering
\begin{adjustbox}{max width=\textwidth}
\begin{footnotesize}
\begin{tabular}{lllllllll}
\toprule
\multirow{2}{*}{\textbf{Model}}     & \multicolumn{2}{c}{\textbf{short}} & \multicolumn{4}{c}{\textbf{long}}  & \multirow{2}{*}{\textbf{\begin{tabular}[c]{@{}l@{}}Perplexity\\ WikiText-103\end{tabular}}} & \multirow{2}{*}{\textbf{\begin{tabular}[c]{@{}l@{}}Forgetting rate\\ TREx\end{tabular}}}              \\
\cmidrule(lr){2-3}\cmidrule(lr){4-7}
                                    & nd              & fd               & nd         & fd         & md         & mu         \\ 
\midrule
random pivot object                             & 45.55                     & 44.26                           & 30.88                          & 29.85                  & 31.02                  & 9.61        & - & -          \\
\hline
OPT-125M-full-context                            & 47.39                     & 36.41                           & 34.97                          & 31.21                  & 19.36                  & 21.96          & 27.68 & 0.0        \\
OPT-350M-full-context                           & 46.24                     & 39.30                           & 23.41                          & 22.83                  & 19.07                  & 9.82         &22.04 & 0.0          \\
OPT-1.3B-full-context                           & 44.79                     & 39.30                           & 42.48                          & 44.21                  & 37.57                  & 19.94          &14.64 & 0.0        \\
OPT-2.7B-full-context                           & 42.19                     & 37.86                           & 34.97                          & 39.30                  & \textbf{40.75}                  & 17.91              &12.47 & 0.0    \\
OPT-IML-1.3B-full-context                        & 50.86                     & 44.79                           & 41.32                          & 43.35                  & 33.81                  & 18.49           &14.68 & 0.0        \\
OPT-IML-MAX-1.3B-full-context                    & 49.71 & 47.68                           & 40.75                          & 45.08                  & 36.70                  & 18.49                 &14.70 & 0.0 \\
\hline
OPT-125M-finetuned + RMT  & \textbf{93.33}                         & 55.60     & 89.65           &   45.49                     & 23.12                       & 72.42          & 11455.11  & 97.4             \\

OPT-125M + MemoryPrompt             & 89.99                  & \textbf{58.84}  & \textbf{92.14}            & \textbf{51.38}                      &  25.83                      &  \textbf{80.69}        & 27.80 & 13.0               \\
OPT-350M + MemoryPrompt             & 86.35                  & 53.0  & 87.05            & 46.82                      &  22.19                      &  74.33  & 22.10 & 10.0\\
                  
\bottomrule
\end{tabular}
\end{footnotesize}
\end{adjustbox}
\caption{Accuracy of the different models on the fact-updating datasets,\protect \footnotemark{} WikiText-103 perplexity and forgetting rate (\%) on TREx facts, which measure catastrophic forgetting. The non-trivial \textit{random pivot object} baseline picks an object appearing in any pivot statement at random.}
\label{tab:results_fact_update}
\end{table*}

\begin{table*}[tb]
\centering
\begin{footnotesize}
\begin{tabular}{llll}
\toprule
\textbf{Model}                      & \textbf{\begin{tabular}[c]{@{}l@{}}Perplexity\\ MSC\end{tabular}} &  \textbf{\begin{tabular}[c]{@{}l@{}}Perplexity\\ WikiText-103\end{tabular}} & \textbf{\begin{tabular}[c]{@{}l@{}}Forgetting rate\\ TREx\end{tabular}} \\
\midrule
OPT-125M (conditioned on last turn) & 23.26      & 27.68 & 0.0         \\
OPT-125M (conditioned on history)   & 20.79       & 27.68 & 0.0          \\
\hline
OPT-125M-finetuned + RMT  & 17.82    & 202.83         &80.0 \\
OPT-125M + MemoryPrompt             & 20.60               & 28.37 & 13.0\\
\bottomrule
\end{tabular}
\end{footnotesize}
\caption{Perplexity on MSC. We also report perplexity on WikiText-103 and forgetting rate (\%) on TREx facts, which measure catastrophic forgetting. }
\label{tab:msc}
\end{table*}

\begin{figure*}[tb]
    \centering
    \includegraphics[width=0.95\textwidth]{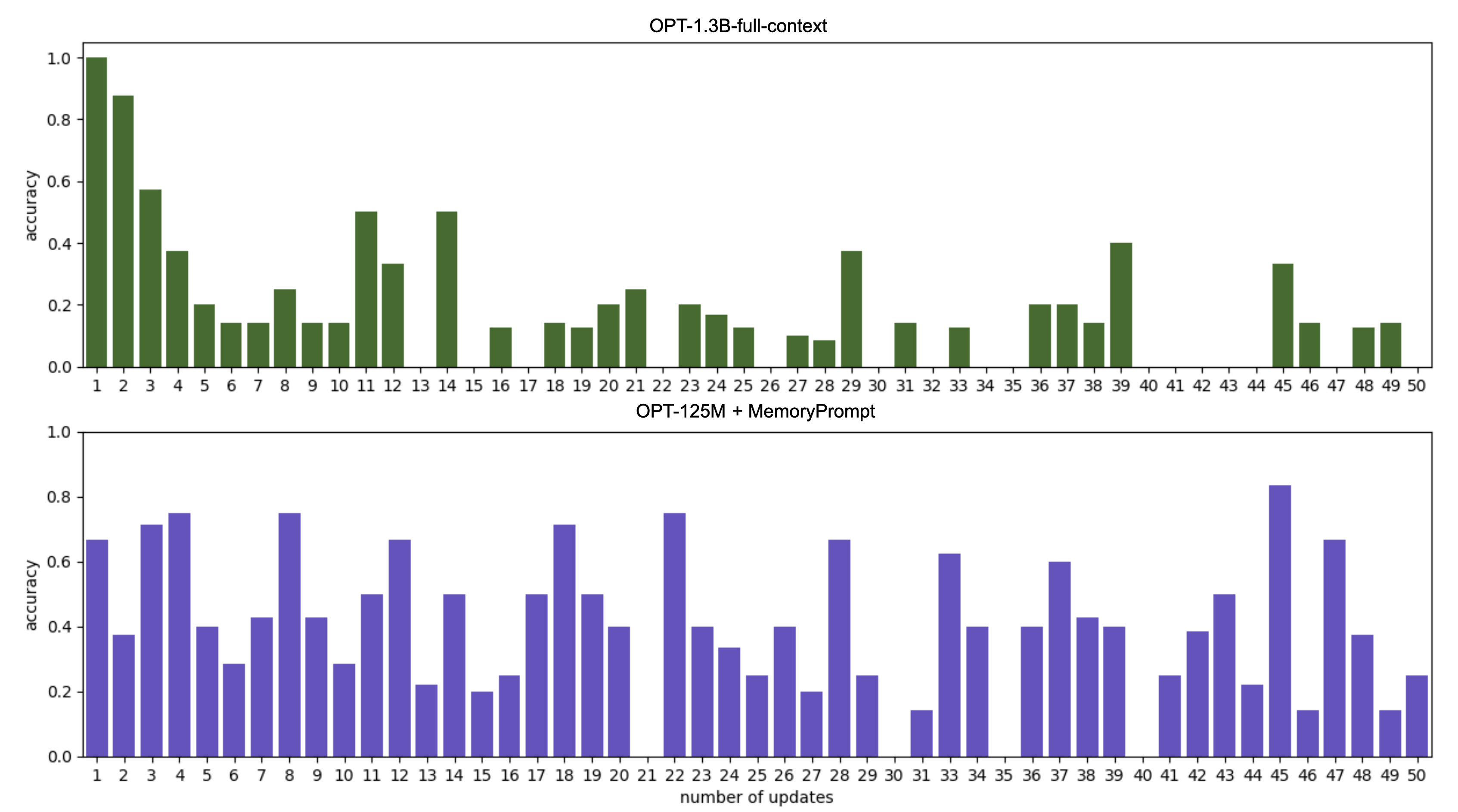}
    \caption{Accuracy of OPT-1.3B full-context (top) and OPT-125M + MemoryPrompt (bottom) as a function of the number of updates on the \textit{long many-updates} (mu) fact-updating dataset.}
    \label{fig:accuracy_updates}
\end{figure*}

\subsection{Fact updating}

Memory-augmented models outperform their full-context counterparts on all the fact-updating datasets (see Table \ref{tab:results_fact_update}). Indeed, MemoryPrompt applied to OPT-125M, the smallest model in the OPT family, outperforms the largest full-context models we were able to run, including instrucion-tuned (\textit{IML}) variants, often dramatically.  

The average input length to the full-context models in the \textit{long} setups is 1,698 tokens, whereas it is only 92 tokens for MemoryPrompt, showing that, for this task, memory vectors are better and much more efficient representations compared to full textual context.

RMT shows good performance overall, but it lags in general behind MemoryPrompt. Importantly, as we show in Section \ref{sec:catastrophic} below, the good performance of RMT comes at the cost of having lost much of the knowledge stored in the underlying model, making this approach unviable for scenarios in which we want a LM to both effectively track contextual updates and maintain its core general knowledge.
 
Surprisingly, OPT-125M outperforms the larger OPT-350M model in both the full-context and memory-augmented scenarios. This might be related to the recent observation by \citet{Voita:etal:2023} that OPT-350M is an ``outlier'' model in the OPT family. In any case, the MemoryPrompt wrapper clearly benefits OPT-350M as well.

As shown in Figure \ref{fig:accuracy_updates}, unlike for the full-context baselines, MemoryPrompt's performance is essentially stable across the number of updates within a dataset. On the other hand, in Table \ref{tab:results_fact_update} we observe a clear slump for all memory-augmented models on the  \textit{long md} dataset, the setting with the largest number of facts to keep track of (recall that models can't tell distractors apart from the pivot, so they must track updates of both). This suggests that MemoryPrompt is robust to the number of updates of specific facts, but it struggles when there are many different facts to update (a problem that RMT also displays). If MemoryPrompt, as suggested by our qualitative analysis in Section \ref{sec:memory-vector-analysis} below, is learning to specialize specific memory vectors to specific facts, it figures that, with just 5 memory vectors and too many facts to keep track of, its performance drops. In other words, the problem is not with the updating and retrieval mechanism, but with the amount of different facts to store, clearly an issue to be pursued in future work.

\footnotetext{The dataset names nd, fd, md and mu come from Table \ref{tab:dataset_config}.}

\subsection{MSC}

MemoryPrompt's performance on MSC is as good as conditioning on the whole conversation history (see Table \ref{tab:msc}).\footnote{We also evaluate MemoryPrompt on the IRC Disentanglement \citep{kummerfeld-etal-2019-large} dialog dataset and obtain similar results as with MSC. However, we could not train RMT on it due to computational limitations} This implies that the memory vectors can effectively track long previous interactions  without requiring a large context window, even when the relevant information in the previous interactions is not as clearly encoded as in the fact updating datasets. We note that the entire history contains 1,412 tokens on average vs.~18 tokens of context for MemoryPrompt. RMT achieves lower perplexity, but, as it requires full-model finetuning, this comes at the cost of catastrophic forgetting, as we will discuss next.

\subsection{(No) catastrophic forgetting}\label{sec:catastrophic}
Our intention with MemoryPrompt is to provide a light way for the model to track contextual information \textit{without affecting the model base performance}. In a realistic scenario, a generalist LM should be able to update information during a conversation, but still retain the knowledge it acquired during pre-training. We thus want to ensure that adding the MemoryPrompt vectors to the LM input is not affecting the model on its base task (next token prediction). To test this, we collect 4 prefixes generated at the end of sequences from each fact-updating dataset (24 prefixes in total). For MSC, we collect prefixes generated at the end of conversations. These prefixes are then combined with the input of the base LM (as we normally do when using MemoryPrompt). We apply the same approach to evaluating RMT for catastrophic forgetting.

We evaluate MemoryPrompt and RMT on two tasks probing how their memory vectors change the model's original behaviour. First, we compute their perplexity on a standard corpus that was not used to train the memory component. In particular, we use  WikiText-103 \citep{stephen2017pointer}. We also compute a \textit{forgetting rate} score, that is, the proportion of (non-updated) TREx facts for which a memory-augmented LM predicts a different object than the same model would without any modification (so, trivially, this measure is 0\% for the unchanged full-context models). We compute  forgetting rate on facts that were not used in the fact updating datasets. 

As the \textit{Perplexity} columns of tables  \ref{tab:results_fact_update} and \ref{tab:msc} show, MemoryPrompt's WikiText-103 perplexities are almost identical to those of the equivalent non-augmented models. The situation is very different for RMT, whose WikiText-103 perplexities suggest that the augmented model has become incapable to perform the standard LM token prediction task. Concerning forgetting rate (last column of tables \ref{tab:results_fact_update} and \ref{tab:msc}), we confirm again that RMT is dramatically affected by the memory-augmentation process, making new predictions for a large majority of facts that were not updated. MemoryPrompt's forgetting rate is also non-negligible, but much lower than that of RMT.

Overall, we confirm that a LM augmented with MemoryPrompt's vectors is not greatly affected in its default performance, so that it can both take contextual updates into account, but continue using the rich general knowledge it acquired during pretraining.

\begin{figure*}[tb]
    \centering
    \includegraphics[width=0.95\textwidth]{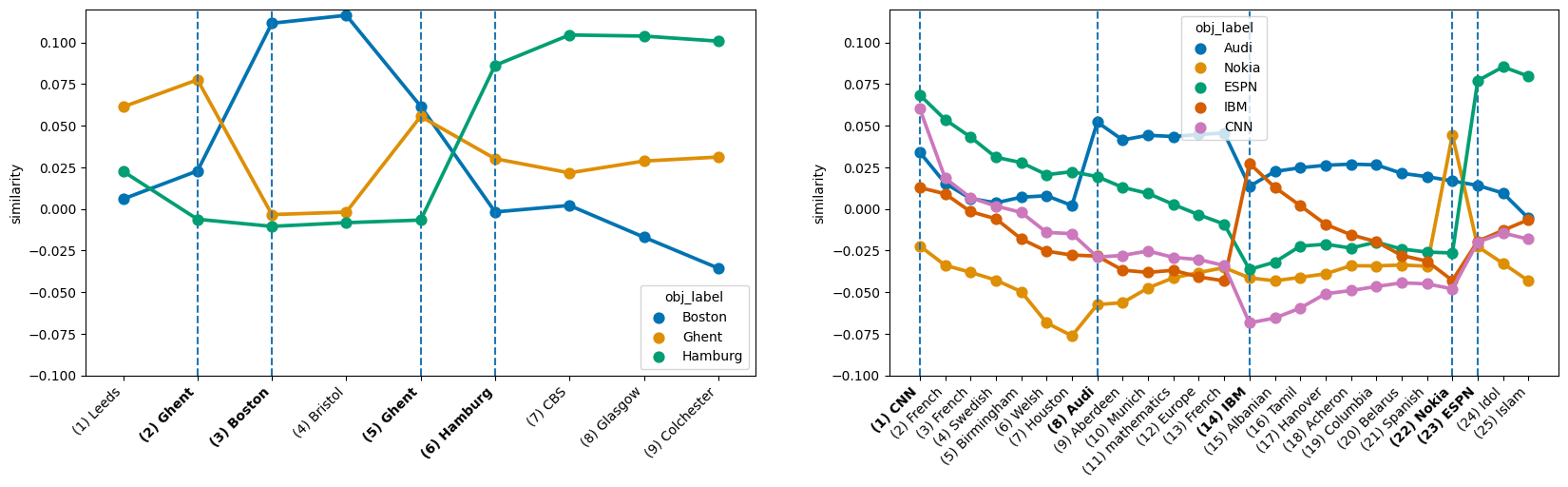}
    \caption{Cosine similarity between one of 5 memory vectors and the embeddings of the objects in a sequence. Each figure represents a sequence, with  objects of different facts on the x-axis. The pivot objects are in bold. The subject/relation of the pivots are \textit{Louis Charles Delescluze}/\textit{work location} (left) and \textit{Rao Remala}/\textit{employer} (right).}
    \label{fig:prompt_analysis}
\end{figure*}

\subsection{Memory vector analysis}\label{sec:memory-vector-analysis}
Memory vectors lie in the same space as the model token embeddings. We can thus get insights on what they are recording by directly measuring their cosine similarity to that of tokens of interests, such as those representing updated objects. Figure \ref{fig:prompt_analysis} shows representative cosine similarity profiles across several updating and non-updating statements for the same memory vector and all the object embeddings (examples taken from the short fact-updating dataset). As the figure shows, this memory vector tends to become more similar to the pivot objects that occur in update statements, and often retains this high-similarity across a number of statements involving other, irrelevant objects. For example, on the left panel, the memory vector increases its similarity to \textit{Ghent}, \textit{Boston}, \textit{Ghent} and \textit{Hamburg}, respectively, as it encounters fact updates involving these objects. Irrelevant facts are ignored, even when they involve cities as objects, which could in principle confuse the model.

This analysis suggests that the MemoryPrompt vectors possess a certain degree of interpretability, although we typically failed to find more than a single memory vector per dataset that possessed this degree of transparency, and we must leave further vector decoding work to future research.

\section{Discussion}

We introduced MemoryPrompt, a simple approach that allows a Transformer-based LM to carry and update contextual information across long spans without requiring a long attention context window. The main intuition of our approach is that relevant information can be passed across processing steps by letting the LM read a set of ``memory'' vectors as a prefix to its standard input, akin to the idea of soft prompting. By piggybacking on the LM standard interface, we can add our memory module to a pre-trained LM without touching its architecture and without fine-tuning the model.

On a dataset designed to test the ability of models to track multiple fact updates, a smaller LM augmented with MemoryPrompt can greatly outperform a much larger LM that gets the whole context as its input. On the long-span dialogue MSC dataset, MemoryPrompt provides comparable performance to that of a model that can access the whole dialogue history from its orders-of-magnitudes larger context window.

We compared MemoryPrompt to our re-implementation of RMT, a state-of-the-art memory-augmented model that serves a similar purpose as MemoryPrompt. While RMT achieves comparable performance in the tasks we considered, it does so at the cost of catastrophic forgetting, that is, erasing the general token-prediction capabilities of the underlying LM. MemoryPrompt is only slightly affected by this problem.

MemoryPrompt still needs to be tested on a more varied set of challenges and applied to larger LMs (our experiments were constrained by computational limitations). Moreover, we have seen that MemoryPrompt is not robust to the number of mutable facts in a sequence, as demonstrated by its performance on the many-distractors dataset (see Table \ref{tab:results_fact_update}). In this setup, OPT-125M + MemoryPrompt outperforms its full-context counterpart, but it lags behind the larger full-context models.

Moving forward, we believe that the light-touch memory tracking abilities of MemoryPrompt have a natural application in adapting pre-trained LMs to specific users, and our main goal for the future is to extend and test the model in a realistic setup in which a LM needs to adapt to a persistent user, by storing user-specific knowledge in its memory. This, in turn, raises interesting questions concerning the nature of memories. Can the memory system, for example, learn which types of facts are user-dependent and highly mutable, and should be constantly tracked and updated? Can our method be extended to track \textit{different} streams of information coming from different interlocutors? We leave these and other questions to future work.

%\section{Limitations}

% In the long-span MSC language modeling task, OPT-125M + MemoryPrompt performs on a par with the same model fed the whole context history (Table \ref{tab:msc}). From a practical point of view, as the model is already available as a pre-trained system with a large context window, there is no practical advantage in using the MemoryPrompt version. In future work, we would like to study training strategies that might allow MemoryPrompt to achieve \textit{lower} perplexity (on this and other datasets) with respect to non-augmented LMs of comparable size or larger (the same result we are already obtaining with the more constrained fact-updating dataset).

% We provided a preliminary analysis of a single memory vector, but we still understand very little about how information is carried through such vectors. For example, since the system does not know \textit{which} statements pertain to pivots, the memory vectors should in principle carry information about all facts. Are they doing this? Do they learn to store different information about mutable and stable facts? We leave these and related questions to future work.

\section{Ethics Statement}

As we are using pre-trained language models without re-training them, we are neither amplifying nor alleviating the ethical issues that come with them. We are also using datasets derived from existing resources, and we are not aware of specific ethical issues pertaining to these resources.

Current state-of-the-art language models are extremely large, and they can only be trained and run by a few companies that have access to enormous computational resources. We see efforts, such as ours, to improve the performance of smaller language models as positively contributing to the need to democratize access to high-performance AI systems.

\section{Acknowledgements}
 We thank Marco Del Tredici and the COLT group for many interesting discussions about the topic of memory in language models. We thank Corentin Kervadec for helpful feedback and suggestions, as well as for help in constructing the fact-updating datasets. Our work was funded by the European Research Council (ERC) under the European Union's Horizon 2020 research and innovation programme (grant agreement No.\ 101019291). This paper reflects the authors' view only, and the funding agency is not responsible for any use that may be made of the information it contains. 

\section{Bibliographical References}\label{sec:reference}
\bibliographystyle{lrec-coling2024-natbib}
\bibliography{custom,marco}

\end{document}